\documentclass[11pt]{article}
\usepackage[letterpaper,margin=0.75in]{geometry}
\usepackage{amsmath,amssymb}
\usepackage{graphicx}
\usepackage{cite}
\usepackage{booktabs}
\usepackage{array}
\usepackage{mathptmx}
\usepackage[hidelinks]{hyperref}
\usepackage{titlesec}
\usepackage{abstract}
\usepackage{authblk}
\usepackage{xcolor}
\usepackage{tabularx}
\usepackage[section]{placeins}

\newcolumntype{L}[1]{>{\raggedright\arraybackslash}p{#1}}

\titleformat{\section}{\normalfont\large\bfseries}{\thesection}{1em}{}
\titleformat{\subsection}{\normalfont\bfseries}{\thesubsection}{1em}{}
\titleformat{\subsubsection}{\normalfont\itshape}{\thesubsubsection}{1em}{}

\title{\LARGE\bfseries From High Recall to High Utility: Dataset-Adaptive Post-Processing of LLM-Generated Customer Intents}
\author{Mahesh Viswanathan, Joan Rossello, Leticia Fernandes, and Paul Mutawe}
\affil{Cisco Systems, Inc., San Jose, CA, USA \\ \texttt{\{mahviswa, joanrb, leticiaf, pmutawe\}@cisco.com}}
\date{}

\begin{document}
\maketitle

\begin{abstract}
\noindent Large language models can extract useful signals from heterogeneous enterprise data, but high-recall extraction often produces outputs that are duplicated, uneven in granularity, semantically overlapping, or too numerous for downstream systems and human reviewers to use effectively. We present a dataset-adaptive post-processing architecture developed in the context of Customer Intent Extraction (CIE), where unstructured customer language is transformed into stable, traceable intent units. The approach deliberately separates recall-oriented extraction from utility-oriented reduction. Source-specific preprocessing first isolates evidence from multimodal plans, sparse operational records, and structured opportunity data. Candidate intents are then standardized and deduplicated, optionally enriched with metadata for embedding computation, represented in a shared semantic vector space, and grouped using a clustering strategy selected according to characteristics of the candidate set. Cluster-level keywords provide an explainability layer, while singleton reassignment requires agreement between embedding and keyword similarity. Finally, constrained language-model aggregation produces one concise intent per cluster without introducing unsupported concepts, and the resulting unit retains provenance, clustering, embedding, and generation metadata. The resulting architecture treats post-processing not as cosmetic cleanup, but as a semantic reduction layer that converts high-recall LLM outputs into reusable enterprise intelligence. Beyond the extraction pipeline itself, we describe two downstream applications of the resulting intent representation: Machine-Generated Intents, which infer likely objectives for customers lacking direct evidence by drawing on peer customers with similar profiles, and intent-guided semantic retrieval and mapping, which uses the stable intent as a query against a downstream decision space, illustrated here by mapping customer intents to business outcomes.

\end{abstract}
\vspace{1em}

\section{Introduction}

Enterprise systems increasingly rely on large language models (LLMs) to recover useful information from documents, operational records, conversations, and other forms of unstructured or semi-structured data~\cite{cheng2025meituan}. In customer-facing settings, however, useful information is rarely expressed as a clean label or a single self-contained sentence. A customer objective may be distributed across presentation tables, free-form notes, key performance indicators (KPIs), opportunity records, next-step fields, and product metadata. Different customers may describe the same objective in different words, while superficially similar statements can refer to materially different needs.

Customer Intent Extraction (CIE) was developed to bridge this gap between natural customer language and the structured representations required by analytics, recommendations, and customer-success workflows. In CIE, a customer intent is an intermediate representation of what the customer is trying to achieve: it combines an overarching business strategy with a specific product, technology, domain, or operational goal. A useful intent must be goal-oriented, grounded in source evidence, appropriately scoped, and reusable beyond the source artifact from which it was inferred.

The central engineering challenge is not simply to prompt an LLM to ``summarize'' customer data. It is to identify what counts as intent, recover the evidence that supports it, preserve context, and consolidate repeated or overlapping statements without erasing distinctions that matter. A raw model response may sound plausible, but a production intent layer must also produce outputs that are stable, traceable, explainable, and structured for downstream consumption.

This work rests on a deliberate asymmetry between extraction and post-processing. The early stages are optimized for \emph{recall}: when the relevant evidence is not known in advance, candidate objectives should be captured broadly enough that important signals are unlikely to be lost. The post-processing stages are optimized for \emph{utility}: the resulting candidate set is standardized, deduplicated, semantically grouped, checked for ambiguity, and reduced to a smaller set of stable intent units. The system therefore accepts redundancy early in the pipeline in exchange for a controlled and auditable reduction later.

The primary contribution of this article is a post-processing architecture for transforming high-recall LLM-generated candidate intents into a smaller set of stable, explainable, and reusable intent units. Specifically, we describe:

\begin{enumerate}
\item A high-recall-to-high-utility design in which broad candidate extraction is followed by a dedicated semantic reduction stage;
\item A dataset-adaptive post-processing pipeline that combines deterministic standardization, optional metadata-aware semantic representation, and clustering strategies selected according to characteristics of the candidate set;
\item A conservative approach to ambiguity and outliers, in which singleton intents are reassigned only when multiple semantic signals provide sufficient evidence;
\item Constrained cluster-level aggregation that reduces redundancy while preserving the distinct information contained in the underlying intents; and
\item A traceable output representation that retains cluster membership, source provenance, semantic descriptors, and model and processing metadata alongside each consolidated intent.
\end{enumerate}

The remainder of the article focuses on the construction of the intent intelligence layer and, in particular, the post-processing architecture. Later sections describe two applications of the resulting intent representation: Machine-Generated Intents (MGI), which infer candidate intents for customers using peer and metadata information, and intent-guided semantic retrieval and mapping, in which an intent is used to retrieve and rank relevant downstream entities. Business-outcome mapping is presented as one concrete implementation of this broader retrieval pattern. Evaluation and the broader engineering implications of this architecture are then discussed as separate concerns.

\section{Literature Review}

This article sits at the intersection of LLM-based information extraction, semantic representation, clustering, and explainable enterprise AI. The technical components used by the system--embedding models, cosine similarity, hierarchical clustering, dimensionality reduction, density-based clustering, keyword generation, and constrained language-model synthesis--are established techniques. The contribution described here is the coordinated use of these components as a reduction layer after deliberately high-recall extraction, with the clustering workflow selected according to characteristics of the candidate set and with uncertainty and provenance preserved explicitly.

\subsection{Intent Extraction and Business Language Understanding}

Extracting actionable intent from natural language has been studied extensively outside the enterprise setting. Corbeil et al.~\cite{corbeil2022intentlandscape} build an intent landscape directly from real-world conversational corpora using extractive question-answering transformers, addressing a problem close to ours: recovering a structured objective space from heterogeneous, freeform customer language rather than from a fixed label set. Manias et al.~\cite{manias2024intentbased} apply large language models to intent extraction in a different domain, translating natural-language network operator intents into structured configuration for 5G core networks; their work shares our motivation of using an LLM to bridge unstructured intent expression and a structured downstream representation, though the target representation there is a deterministic network configuration rather than a reusable semantic unit. Liu et al.~\cite{liu2025adretrieval} infer commercial intention from user behavior for sponsored search ad retrieval, illustrating a parallel high-volume, business-facing setting in which inferred intent becomes the input to a retrieval pipeline, similar in spirit to our intent-guided semantic retrieval and mapping application. At the systems level, Cheng et al.~\cite{cheng2025meituan} describe how large language models were deployed across Meituan's customer interaction systems to jointly improve customer satisfaction and reduce operating cost, reinforcing the broader premise of this article: production customer-facing LLM systems succeed or fail on data engineering, representation, and evaluation discipline rather than on model choice alone. None of these works, however, treat the extracted intent itself as a versioned, provenance-carrying data asset that is deliberately over-generated and then reduced; that framing is the specific contribution of CIE's post-processing layer.

\subsection{Semantic Representation, Clustering, and Deduplication}

The embedding, clustering, and deduplication components used in the post-processing pipeline build on established techniques. Sentence-level semantic representations are computed using the Sentence-BERT family of siamese encoders~\cite{reimers2019sentencebert}, which made cosine-similarity comparison between short texts computationally practical at the scale required by CIE's candidate-intent volumes. For larger candidate sets, the adaptive clustering policy uses Uniform Manifold Approximation and Projection (UMAP)~\cite{mcinnes2018umap} for dimensionality reduction followed by Hierarchical Density-Based Spatial Clustering of Applications with Noise (HDBSCAN)-style density-based clustering~\cite{mcinnes2017hdbscan}, chosen because, unlike centroid-based methods, it tolerates clusters of uneven size and density and exposes outliers explicitly rather than forcing every point into a group. The standardization stage's exact-deduplication logic is a deterministic special case of a broader problem: semantic deduplication of large, noisy text collections, studied by Abbas et al.~\cite{abbas2023semdedup} in the context of curating web-scale training corpora. CIE's contribution here is not any one of these components but their coordinated, dataset-adaptive application as a post-extraction reduction layer, with the selection between clustering families made explicit and recorded rather than fixed in advance.

\subsection{Constrained Generation, Faithfulness, and Evaluation}

Because both the keyword-extraction and constrained-aggregation stages rely on an LLM, the pipeline inherits the LLM's well-documented tendency to introduce unsupported content. Huang et al.~\cite{huang2023hallucination} provide a taxonomy of hallucination in large language models that motivates the explicit factual constraint placed on the aggregation generator: aggregation is restricted to information already present in the cluster's member intents rather than treated as open-ended synthesis. On the evaluation side, using a capable LLM as an automatic judge of open-ended generation quality follows the practice formalized by Zheng et al.~\cite{zheng2023judging}; this article treats that practice as one input among several, alongside subject-matter expert (SME) review, rather than as a sufficient validation step on its own.

\subsection{Retrieval-Based Intent-to-Outcome Mapping and Predictive Intent Generation}

The intent-guided semantic retrieval and mapping application uses a bi-encoder for high-recall candidate retrieval followed by a cross-encoder for precision re-ranking. This retrieve-fast, rerank-smart pattern is the same one studied by Geigle et al.~\cite{geigle2022retrievererank} for cross-modal retrieval, and it generalizes the same architecture introduced for sentence encoders in~\cite{reimers2019sentencebert}: an efficient, independently computed representation narrows the candidate space before a more expensive, jointly computed representation refines the final ranking. Machine-Generated Intents, which infer likely objectives for a customer from peer customers' known intents, is a content-based instance of the cold-start recommendation problem: Viljanen et al.~\cite{viljanen2020coldstart}, working in a game-recommendation setting, address the same underlying challenge of producing useful recommendations for entities with no direct interaction history by relying on profile-level feature similarity, which is the same mechanism MGI applies at the customer level. Finally, this work extends the authors' broader interest in coordinating multiple AI components into auditable, production-grade enterprise systems: DAWN~\cite{aminiranjbar2025dawn}, the authors' prior work on a framework for orchestrating distributed AI agents, shares with the architecture described here the practice of wrapping every model invocation in an explicit, versioned, and inspectable structure so that a system built from stochastic components can still produce outputs an enterprise can trust, trace, and reprocess.

\section{Problem Definition and Customer Intent}

\subsection{Definition}

CIE treats intent as a reusable data asset rather than a one-time model output. An intent is a statement that represents the customer's strategic objective and the specific goal associated with a product, technology, domain, or operational context. It is not equivalent to an activity, a next step, or a product mention.

For example, ``\emph{uploaded the latest architecture deck}'' describes a completed activity, while ``\emph{run an adoption workshop}'' describes a next step. ``\emph{Catalyst Center deployment}'' identifies a topic or technology. By contrast, ``\emph{improve network visibility to identify performance issues earlier}'' expresses a desired operational outcome with enough context to be actionable as an intent. A statement such as ``\emph{improve visibility}'' may be a candidate intent but remains underspecified until product, domain, or other evidence is available.

This distinction matters because the same lexical surface form may hide different objectives. Visibility in network operations can refer to performance monitoring and faster troubleshooting, whereas visibility in a security context can refer to threat detection, user activity, or compliance reporting. Customer intent extraction is therefore not adequately described as a fixed-label classification problem.

\subsection{Design desiderata}

A stable intent unit is designed around seven properties: \textbf{Faithful} (the final statement contains only concepts supported by member intents and source evidence); \textbf{Traceable} (the raw candidates and source references remain available); \textbf{Explainable} (cluster keywords, representative members, centroids, and clustering metadata reveal how the output was formed); \textbf{Consistent} (canonical metadata and constrained generation reduce avoidable variation); \textbf{Uncertainty-aware} (unusual or weakly supported candidates can remain independent rather than being forced into a group); \textbf{Versioned} (model, prompt, embedding, and clustering configurations are recorded); and \textbf{Product-consumable} (the representation is predictable enough to support downstream systems).

\begin{figure}[htb]
\centering
\includegraphics[width=1\textwidth]{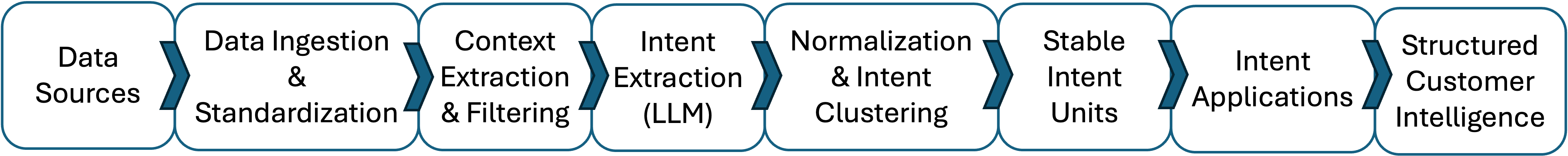}
\caption{End-to-end operating model. Source-specific preparation preserves evidence, while a common post-processing layer transforms high-recall candidate intents into stable outputs.}
\label{fig:pipeline}
\end{figure}

Figure~\ref{fig:pipeline} shows the end-to-end operating model. Stability does not imply immutable wording. Models, prompts, thresholds, and taxonomies will evolve. Stability means that downstream systems receive a consistent semantic object with enough provenance to explain, compare, and reprocess the representation as the implementation changes.

\section{Intent Extraction}

\subsection{Why recall is favored early}

When the relevant evidence is unknown in advance, optimizing the first inference step for apparent precision can silently discard useful objectives. Customer evidence may appear in strategic-priority tables, initiative descriptions, KPIs, operational notes, or multiple fragments across a document. CIE therefore separates two objectives that are often combined: an extraction stage that maximizes evidence and candidate coverage, and a post-processing stage that maximizes usability and reduces semantic redundancy.

The cost of this choice is predictable. The raw candidate set tends to contain exact duplicates, paraphrases, uneven levels of granularity, vague statements, and repeated evidence from different sources. The post-processing layer is designed specifically to absorb this cost.

\subsection{Source-specific context preparation}

The preprocessing problem differs substantially by source. Applying one generic preprocessing strategy to every source would either remove useful evidence or pass excessive noise through to the LLM.

\subsubsection{End-to-End plans}

End-to-End (E2E) plans are multimodal presentation documents in which meaning is often carried by layout as much as by text. Tables, merged cells, headings, and spatial relationships can determine which customer priority is associated with which initiative or product. SMEs identify target chapters containing high-value intent evidence. A PowerPoint (PPTX) parser is used to extract corresponding table content into structured JSON, followed by manual validation. This produces a reference representation against which multimodal extraction can be assessed, including keys, cell values, table structure, and merged-cell behavior.

The objective at this stage is not to write the intent. It is to produce a compact and faithful evidence package that allows the intent extractor to reason over the relevant customer context without carrying the entire source presentation into the prompt.

\subsubsection{Customer Success Console}

Customer Success Console (CS Console) data presents the opposite problem: the source is already fielded, but it is extremely sparse and repetitive. Many records do not include an in-scope product, free-form fields can be operationally uninformative, supporting fields are frequently empty, and exact or near-exact duplicates are common.

\begin{figure}[htb]
\centering
\includegraphics[width=0.6\textwidth]{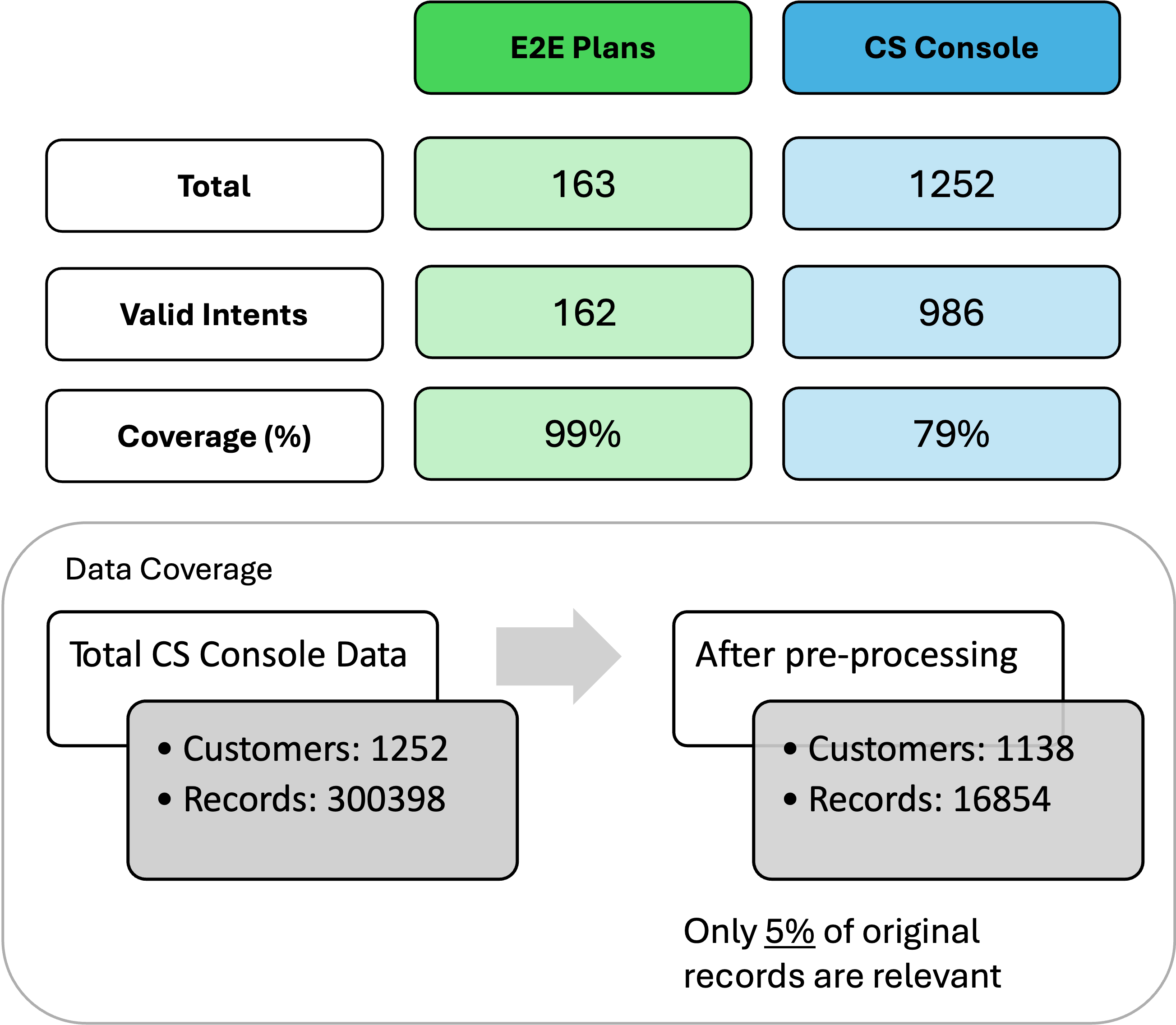}
\caption{Data reduction and coverage for E2E Plans and CS Console sources. For each source, the chart reports the total number of raw records considered, the number retained as valid candidate intents after preprocessing, and the resulting coverage percentage, illustrating that aggressive filtering removes the large majority of raw records while preserving most customer coverage.}
\label{fig:data}
\end{figure}

The preprocessing pipeline therefore applies deterministic filters before LLM inference: records are restricted to non-empty in-scope product values, exact duplicates are removed, and regex rules reject known low-signal patterns such as meeting counts, document uploads, and similar administrative updates. In the dataset used during development, this reduced 300{,}398 records from 1{,}252 customers to 16{,}854 records from 1{,}138 customers, i.e., roughly five to six percent of the original records while preserving most customer coverage, as shown in Figure~\ref{fig:data}.

\begin{table}[htb]
\centering
\caption{Source-specific preparation before common intent extraction.}
\label{tab:sources}
\begin{tabular}{L{2.8cm}L{4.4cm}L{4.4cm}L{4.4cm}}
\toprule
\textbf{Dimension} & \textbf{E2E plans} & \textbf{CS Console} & \textbf{Sales Data Sources} \\
\midrule
Primary form & Slides, tables, merged cells, structured narrative & Large sets of fielded records and free-form entries & Related structured tables containing intent, opportunity, and product information \\
Main failure mode & Loss of layout and table relationships & Sparse fields, low-signal updates, duplicates, missing product context & Relevant context distributed across tables and inconsistent product representations \\
Core preparation & Target-chapter selection, multimodal table extraction, structured validation & Product-scope filtering, regex rejection, empty-field handling, exact deduplication & Table joins, product mapping, feature engineering, schema normalization \\
Output & Compact structured context package & Filtered records grouped by customer and product & Structured intent/opportunity record with enriched product context \\
\bottomrule
\end{tabular}
\end{table}

\subsubsection{Structured opportunity sources}

Table~\ref{tab:sources} summarizes these differences across sources. Structured sales data sources require a different form of preparation: the challenge is less about recovering presentation layout or filtering a single sparse table and more about assembling related information distributed across multiple datasets. Sales data source preparation, for example, combines an intent source table, opportunity line items, and product hierarchy information. Feature engineering maps product context to each candidate and produces a structured record that can be normalized into the common CIE schema.

\subsection{Raw Intent Extraction and Representation}

Once relevant context has been isolated, a source-aware LLM extractor generates candidate intents. Different prompt patterns are used for sources with different evidence structure. E2E prompts operate on structured plan context, while CS Console prompts operate on filtered record context and can include few-shot examples that demonstrate the expected intent style.

The extractor is instructed to recover the customer objective supported by the evidence rather than summarize every detail. Its output is explicitly treated as a candidate, not a final intent. The same underlying goal can therefore appear multiple times when supported by different records or sources.

Each candidate preserves both its generated statement and its provenance. A practical raw intent object contains the intent text, source identifiers, trace pointers, a model-predicted product value, a canonical product value after standardization, customer or account scope, and generation metadata such as prompt version, model version, and timestamp. This provenance becomes essential during deduplication, debugging, reprocessing, and evaluation. Formally, let a candidate set be
\begin{equation}
I = \{(x_i, m_i)\}_{i=1}^{N}, \label{eq:candset}
\end{equation}
where $x_i$ is the candidate intent text and $m_i$ is a metadata object containing source, product, customer scope, and generation information. The post-processing layer maps $I$ to a smaller collection of clusters
\begin{equation}
C = \{C_1, \ldots, C_K\}, \quad K \le N, \label{eq:clusters}
\end{equation}
and generates one consolidated intent $a_k$ for each final cluster $C_k$.

\section{Dataset-Adaptive Post-Processing}

The post-processing pipeline consists of eight logical steps: standardization, context padding, embedding generation, clustering strategy selection, clustering, keyword extraction and normalization, outlier handling, and constrained aggregation. These stages are intentionally modular so that individual components can be evaluated, tuned, or replaced without changing the output contract.

\subsection{Standardization and exact deduplication}

Standardization creates a comparable input set before any semantic operation. Whitespace and casing are normalized, unusable or empty items are rejected, and exact duplicate text is collapsed~\cite{abbas2023semdedup}. When duplicate strings carry different source or product predictions, the records are merged rather than discarded so that provenance is preserved.

\begin{table}[htb]
\centering
\caption{Post-processing stages and their engineering purpose.}
\label{tab:stages}
\begin{tabular}{L{1cm}L{3.1cm}L{5.8cm}L{5.8cm}}
\toprule
\textbf{Step} & \textbf{Stage} & \textbf{Purpose} & \textbf{Key output} \\
\midrule
1 & Standardization & Remove formatting variation, exact duplicates, empty items, and metadata inconsistency & Clean candidate intent objects \\
2--3 & Context padding and embeddings & Enrich short/ambiguous text for embedding only and map candidates to a common semantic vector space & Padded representation, unchanged display text, embedding and model version \\
4--5 & Adaptive clustering & Select and apply an interpretable/scalable grouping strategy according to candidate-set characteristics & Cluster assignments, centroids, method and parameters \\
6 & Keyword extraction & Describe shared cluster meaning and provide an explainability signal & Normalized cluster keyword set \\
7 & Outlier handling & Preserve unique intents or cautiously repair uncertain singleton assignments & Updated clusters plus retained singletons \\
8 & Constrained aggregation & Produce one faithful, non-redundant statement per cluster & Combined intent plus generation metadata \\
\bottomrule
\end{tabular}
\end{table}

Table~\ref{tab:stages} summarizes all eight stages. Product metadata is similarly mapped to approved canonical keys. Variants, abbreviations, or model predictions can be reconciled using maintained mappings or semantic similarity against known names and aliases. The original predictions are still retained for debugging.

This stage is deliberately deterministic. Exact problems should be solved with exact logic: an LLM is not required to trim whitespace, detect identical strings, or enforce a known product mapping. Reserving probabilistic methods for genuinely semantic decisions reduces cost and improves interpretability.

\subsection{Context padding and semantic representation}

Short intents can be semantically unstable in isolation. For example, ``improve visibility'' may refer to application performance, network operations, threat detection, user activity, or compliance. For small or semantically sparse candidate sets, the system can create an embedding-only representation by appending selected metadata such as product, topic, domain, or customer scope.

Let $p_i$ denote the text used for embedding. If a padding condition is satisfied, $p_i = x_i \oplus g(m_i)$; otherwise, $p_i = x_i$, where $g(m_i)$ serializes only the selected contextual metadata and $\oplus$ denotes concatenation. Crucially, $p_i$ does not replace the original customer-facing text $x_i$. The padded representation is an internal semantic scaffold used only for embedding computation.

A common embedding model $f_\theta$ is then applied:
\begin{equation}
e_i = f_\theta(p_i). \label{eq:embed}
\end{equation}
Semantic proximity is measured using cosine similarity,
\begin{equation}
s(i,j) = \frac{e_i^{\top} e_j}{\lVert e_i \rVert \lVert e_j \rVert}. \label{eq:cosine}
\end{equation}
The embedding model and version are stored because changes to the embedding space can alter similarity distributions and cluster boundaries.

\subsection{Adaptive clustering rather than one algorithm for every scale}

The number of candidate intents can vary substantially by customer, product, and source. A clustering method that is transparent for ten statements can be inefficient or unstable for hundreds. The initial CIE implementation therefore selects among materially different clustering families according to candidate-set size.

\begin{figure}[htb]
\centering
\includegraphics[width=0.85\textwidth]{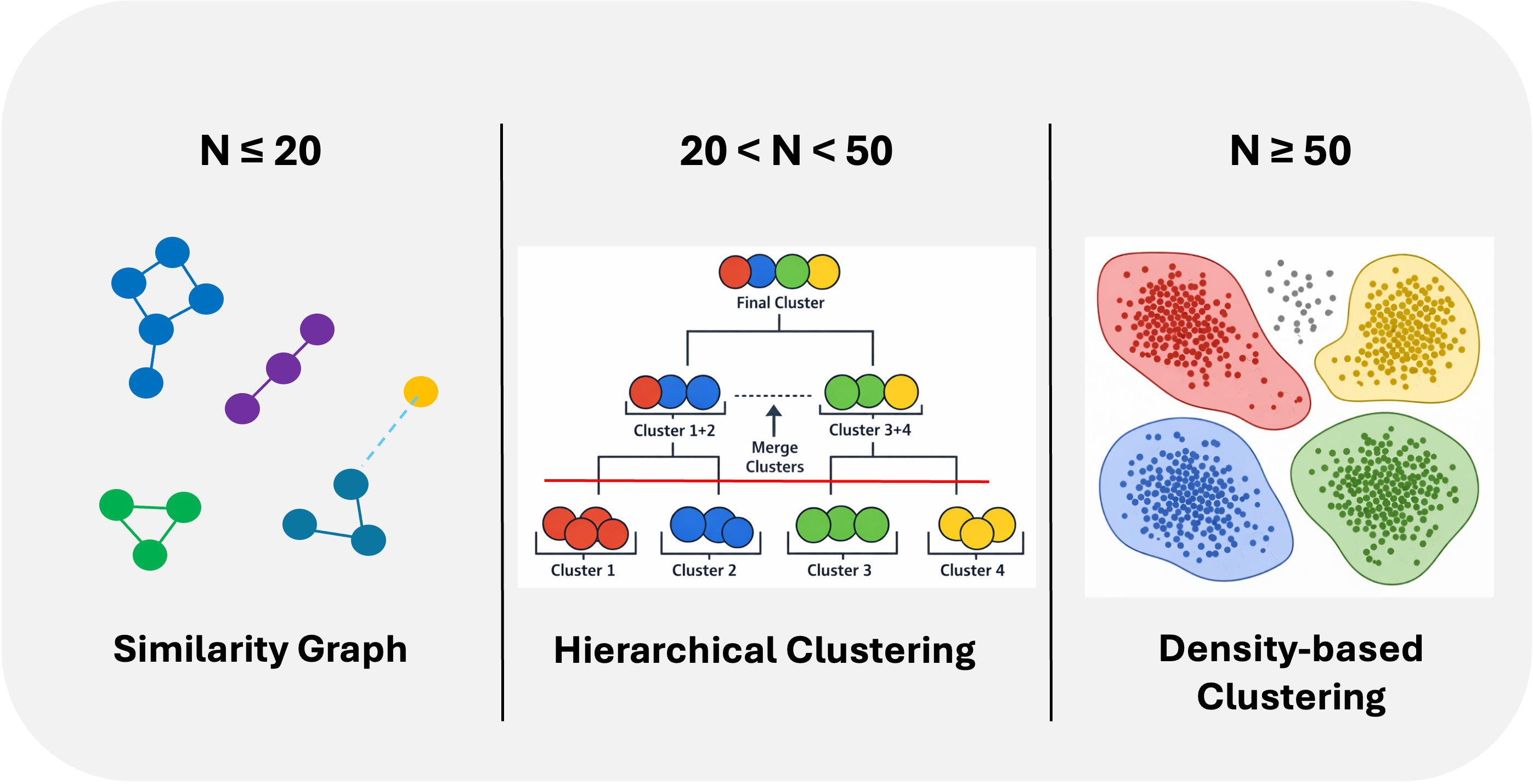}
\caption{Dataset-size-based clustering policy used in the initial implementation. Thresholds are tunable.}
\label{fig:clustering}
\end{figure}

The operational policy, shown in Figure~\ref{fig:clustering}, is that $G_{\text{method}}(N)$ is a pairwise similarity graph when $N \le 20$, hierarchical agglomerative clustering when $20 < N < 50$, and dimensionality reduction followed by density-based clustering when $N \ge 50$.

These values are engineering starting points, not universal scientific constants. They are configurable and should be tuned using historical data, cluster purity, singleton rate, total cluster count, runtime, and human review quality. The central idea is the adaptive selection itself, together with recording the selected method and parameters for reproducibility.

For small sets, pairwise cosine similarities are computed and edges are created when similarity exceeds a calibrated threshold. Connected components become clusters. The method is highly inspectable but can be sensitive to threshold choice and single-link chaining.

For medium-sized sets, hierarchical agglomerative clustering provides a global view of group structure while remaining relatively easy to inspect. Cosine distance and a linkage rule construct a dendrogram, which is partitioned using a tuned distance threshold.

For larger sets, the implementation uses dimensionality reduction via UMAP~\cite{mcinnes2018umap} followed by density-based clustering (HDBSCAN)~\cite{mcinnes2017hdbscan}. This approach can identify clusters of different sizes and densities while also flagging isolated intents as potential outliers, making it more suitable for larger candidate sets.

\subsection{Cluster-level keywords as an explainability layer}

After clustering, an LLM generates semantic descriptors for the shared themes in each cluster. Singleton clusters receive keywords from their individual statement. Raw keywords are normalized for casing and punctuation, deduplicated, and stripped of irrelevant tokens; domain synonyms can optionally be merged using maintained dictionaries or controlled AI assistance.

Keywords are not a replacement for the combined intent. They serve three roles. First, they allow a reviewer to inspect why a group of statements was considered coherent. Second, they provide an independent semantic signal during outlier handling. Third, they give the aggregation step an explicit set of concepts to preserve.

Keyword quality can be evaluated along at least two dimensions: \emph{coverage}, or whether the descriptors capture the content of the member intents; and \emph{distinctiveness}, or whether neighboring clusters receive meaningfully different descriptors.

\subsection{Outlier handling and visible uncertainty}

Clustering creates ambiguity as well as structure. Treating every singleton as an error can discard legitimate low-frequency customer objectives, whereas forcing every item into a cluster creates false certainty.

\begin{figure}[htb]
\centering
\includegraphics[width=0.92\textwidth]{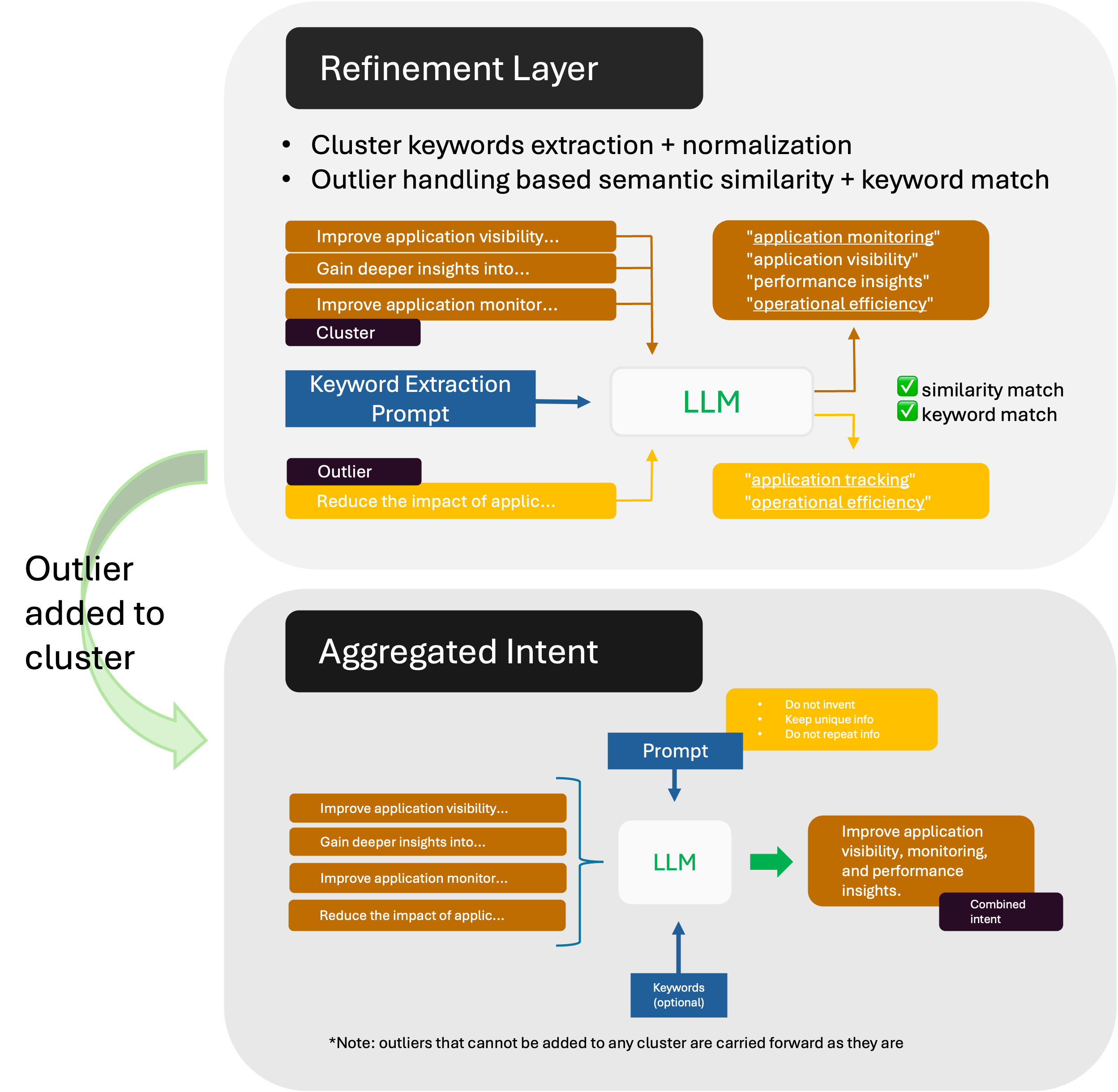}
\caption{Keyword extraction, outlier handling and constrained intent aggregation in the post-processing pipeline.}
\label{fig:refinement}
\end{figure}

CIE therefore uses a conservative reassignment policy for singleton intents, illustrated in Figure~\ref{fig:refinement}. Let $u$ denote a singleton intent and let $C_k$ denote a candidate destination cluster. Each intent $i$ in $C_k$ has an embedding vector $e_i$. The centroid of the cluster, $c_k$, is calculated as the mean of the embeddings of its member intents:
\begin{equation}
c_k = \frac{1}{|C_k|} \sum_{i \in C_k} e_i. \label{eq:centroid}
\end{equation}
The semantic similarity between the singleton and the cluster is then measured using cosine similarity between the singleton embedding $e_u$ and the cluster centroid:
\begin{equation}
s_e(u,k) = \cos(e_u, c_k), \label{eq:sim_e}
\end{equation}
where $s_e(u,k)$ is the embedding-similarity score. A second score, $s_w(u,k)$, measures semantic similarity between the keywords associated with the singleton and those associated with the cluster.

The singleton is reassigned to cluster $C_k$ only when both scores exceed their respective thresholds:
\begin{equation}
s_e(u,k) \ge \tau_e \ \wedge\ s_w(u,k) \ge \tau_w, \label{eq:threshold}
\end{equation}
where $\tau_e$ and $\tau_w$ are the minimum embedding- and keyword-similarity thresholds. If several clusters satisfy both conditions, the strongest-supported cluster is selected. If none satisfy them, the singleton remains as an independent intent.

The use of two signals is intentionally conservative. The system accepts that the correct representation may be ``unique or uncertain'' rather than manufacturing a false match. Outlier rate, reassignment rate, and human acceptance of reassigned examples are useful signals for threshold calibration.

\subsection{Constrained aggregation}

The final step converts each cluster into one concise statement. A language model receives the cluster members and, optionally, the normalized keywords. The generation instruction is narrow: remove repetition, preserve unique supported concepts, and do not introduce information not present in the cluster~\cite{huang2023hallucination}.

For example, a cluster containing ``improve application visibility and monitoring for operations,'' ``gain deeper insights into application performance,'' and ``improve application monitoring for operational efficiency'' can be aggregated into a statement such as ``improve application visibility, monitoring, operational efficiency, and performance insights.'' The objective is aggregation rather than creative summarization.

If $K_k$ is the normalized keyword set for cluster $C_k$, the final representation can be written abstractly as
\begin{equation}
a_k = G_\phi(C_k, K_k; R), \label{eq:aggregation}
\end{equation}
where $G_\phi$ is the generation model and $R$ is a set of factual constraints requiring the output to remain within the information supported by $C_k$.

\begin{figure}[htb]
\centering
\includegraphics[width=0.92\textwidth]{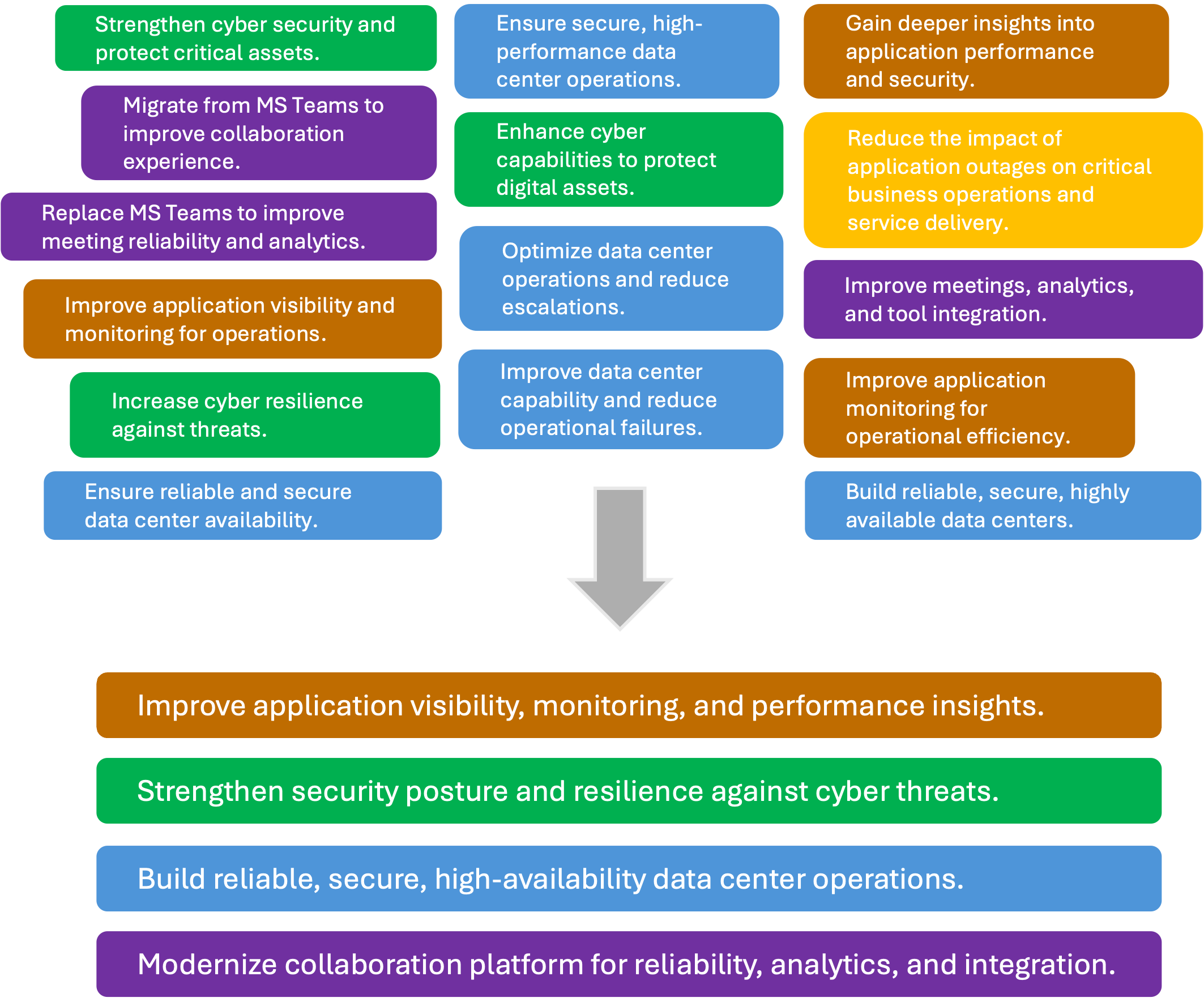}
\caption{Illustrative example of the recall-utility operating model. The color coding shows the aggregation of similar intent units and an outlier (yellow) added to an existing cluster (brown).}
\label{fig:illustrative}
\end{figure}

\subsection{The Stable Intent Unit}

The output of post-processing is more than a sentence. Each cluster is stored as a traceable semantic object containing: a cluster identifier; the cluster centroid; a reference intent, typically a member closest to the centroid; the complete member-intent list; normalized keywords; the canonical product; the combined intent; the clustering method and parameters; embedding and LLM versions; and source references.

This schema insulates downstream consumers from model-level implementation changes. A retrieval or recommendation system can consume the stable combined intent, while investigators can inspect the source candidates, keywords, clustering method, and model versions when a result needs to be explained or reproduced.

\subsection{Illustrative Example}

A worked example from the post-processing design starts with 15 free-text candidate insights spanning cybersecurity, collaboration, application monitoring, data-center operations, and one application-outage statement. Because $N = 15$, the adaptive policy selects pairwise similarity clustering.

The initial clustering produces four multi-member groups and one singleton. The groups correspond to cybersecurity, collaboration platform modernization, application monitoring, and data-center operations. The singleton describes reducing the impact of application outages on business operations and service delivery.

Cluster-level keywords are then generated. The application-monitoring cluster, for example, is described by terms such as application visibility, application monitoring, performance insights, operational efficiency, and application security. The singleton is described by application outages, service delivery, business operations, and operational resilience.

The singleton is compared against each existing cluster using both centroid similarity and keyword similarity. In the scoring process, the application-monitoring cluster provides the strongest agreement (embedding similarity 0.84 and keyword similarity 0.79), while the remaining clusters are substantially weaker. The singleton is therefore reassigned to that cluster, as shown in Figure~\ref{fig:illustrative}.

Constrained aggregation then produces four final intent statements: (1) improve application visibility, monitoring, and performance insights; (2) strengthen security posture and resilience against cyber threats; (3) build reliable, secure, high-availability data-center operations; and (4) modernize the collaboration platform for reliability, analytics, and integration.

This example reduces 15 overlapping and variably worded statements to four explainable outputs while retaining the substantive themes of the candidate set. It is intended to illustrate the mechanics of the method rather than serve as a quantitative benchmark.

\section{Applications of the Intent Intelligence Layer}

Once customer objectives have been represented as stable intent units, the representation can be reused beyond the extraction pipeline itself. In CIE, this enables a shift from identifying what a customer has explicitly expressed to using the accumulated intent corpus as a reusable knowledge layer. Two applications illustrate this transition: Machine-Generated Intents, which infer plausible objectives where direct evidence is sparse, and intent-guided semantic retrieval and mapping, which uses an intent to identify relevant entities in a downstream decision space.

\subsection{Machine-Generated Intents}

Machine-Generated Intents (MGI) extend intent coverage to customers for whom direct source evidence is unavailable or incomplete, but for whom sufficient customer metadata and peer information exist. Instead of extracting an objective from an E2E plan, CS Console record, or another source artifact, MGI uses the intents already observed for comparable customers as evidence from which candidate objectives can be inferred. This makes MGI a content-based instance of the cold-start recommendation problem~\cite{viljanen2020coldstart}, in which direct interaction or intent history is missing and profile-level information is used instead to identify relevant evidence from better-observed entities.

\begin{figure}[htb]
\centering
\includegraphics[width=0.92\textwidth]{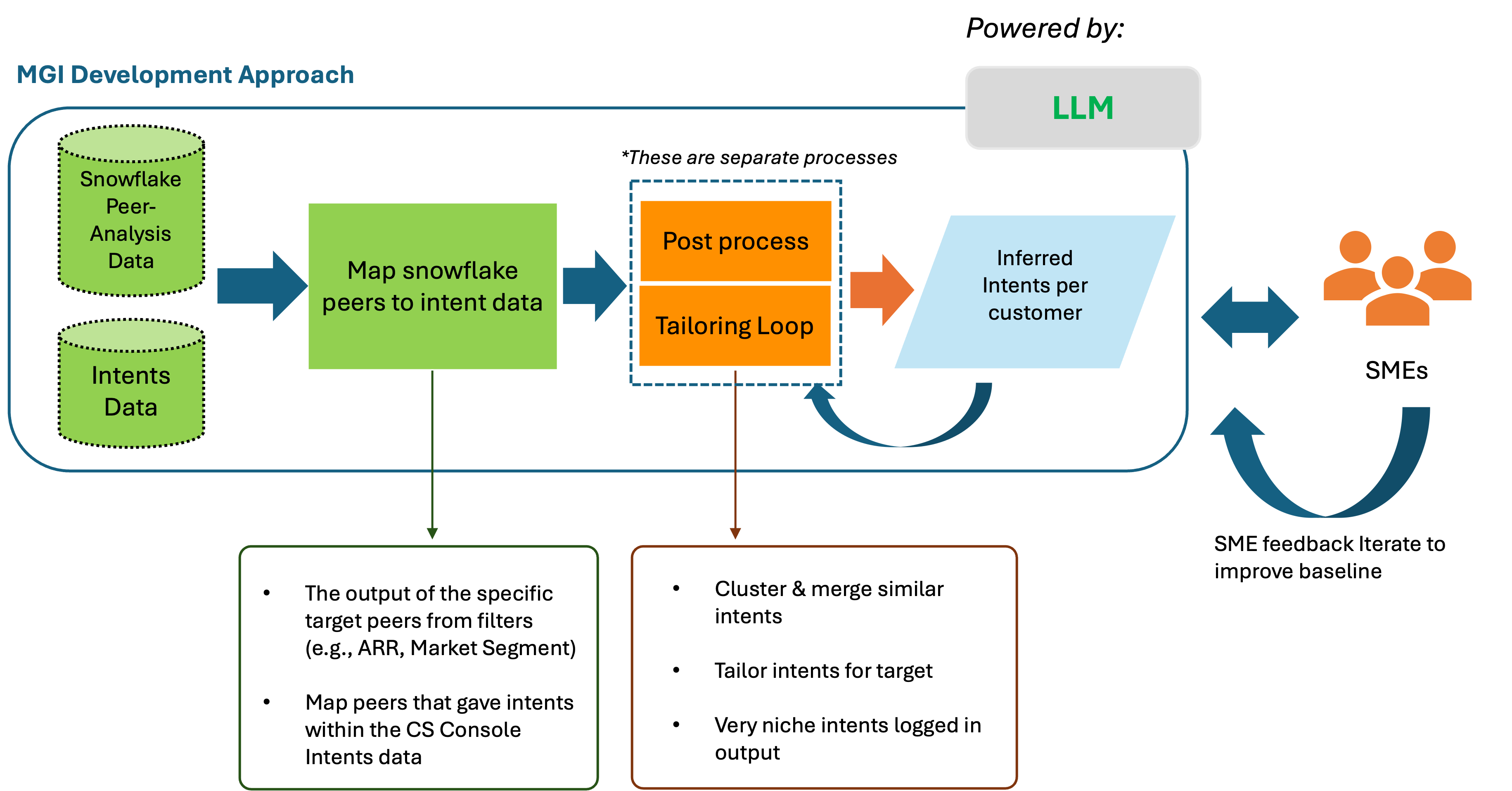}
\caption{Current development flow for Machine-Generated Intents. Deterministic peer-selection criteria identify relevant donor customers, whose observed intents are subsequently post-processed and tailored to the target customer. SME feedback provides an iterative validation loop for both peer relevance and inferred intents.}
\label{fig:mgi-pipeline}
\end{figure}

The current development flow deliberately separates \emph{peer selection} from \emph{intent inference}. Peer identification is treated as a transparent retrieval problem based on business attributes rather than as a learned semantic-neighbor problem: SME-defined rules select donor customers through deterministic filters and ranges--for example, matching market segment while constraining annual recurring revenue to an appropriate range--with attributes such as industry, geography, and customer segment available as additional context. This makes the peer-selection stage directly inspectable and allows the business criteria defining a relevant peer group to evolve independently of the semantic intent-processing models. The resulting peers are then mapped to the existing intent corpus so that only customers with known intent evidence contribute candidate statements. Figure~\ref{fig:mgi-pipeline} summarizes this flow.

Once donor customers have been selected, their known intents pass through a semantic reduction stage. Related candidate intents are clustered and merged, while distinct or weakly supported objectives can remain visible rather than being forced into a dominant theme. The resulting representations are then tailored using metadata belonging to the \emph{target} customer rather than the donor customers. Candidate importance can additionally be informed by the amount of peer support behind a semantic intent--for example, whether an equivalent objective occurs across several selected peers--providing a natural signal for prioritization without treating frequency alone as evidence of correctness.

This separation is important architecturally. Deterministic business rules answer the relatively explainable question of \emph{which customers constitute useful evidence}, while semantic clustering and constrained generation address the more ambiguous question of \emph{what that evidence implies for the target customer}. SME feedback can therefore operate at both levels: reviewing whether the peer-selection rules produce sensible donor populations and whether the inferred intents remain plausible, relevant, and appropriately tailored. More broadly, MGI demonstrates how a normalized intent corpus can become a knowledge source rather than merely the output of an extraction system.

\subsection{Intent-Guided Semantic Retrieval and Mapping}

A second application treats the stable intent as an intermediate representation between a user's expressed goal and the actions, resources, products, services, or other entities that may help satisfy it. Once intent has been represented consistently, it can serve as a semantic query against a structured downstream decision space: a home-buying intent could be matched to properties, financing options, or services; a commercial intent could be matched to products, offers, or next-best actions; and an enterprise customer intent could be mapped to capabilities, adoption objectives, or business outcomes. Intent therefore becomes more than an extracted signal: it provides a reusable semantic interface between unstructured human goals and the structured actions available to an organization. We refer to this broader pattern as intent-guided semantic retrieval and mapping.

Related patterns appear in other domains. Manias et al.~\cite{manias2024intentbased} translate natural-language network intent into structured configuration, while Liu et al.~\cite{liu2025adretrieval} use inferred commercial intention to drive ad retrieval at web scale. These examples illustrate that the downstream target need not be another piece of natural language: once intent has been normalized into a stable semantic object, it can provide the query layer for a range of structured decision spaces.

Within CIE, intent-to-business-outcome mapping is one concrete implementation. Each customer intent is associated with product context, and the candidate outcome space is first restricted to outcomes defined for that product. This deterministic scoping reduces the retrieval space, avoids cross-product false positives, and prevents the semantic models from being asked to distinguish candidates that are operationally irrelevant to the intent.

Retrieval then follows a two-stage bi-encoder/cross-encoder architecture.
A bi-encoder embeds intents and candidate outcomes independently into a shared
semantic space and uses cosine similarity to retrieve a high-recall candidate
set. Because outcome representations can be computed once and reused, this
stage provides efficient retrieval over a comparatively large candidate space.
A cross-encoder subsequently evaluates only the shortlisted intent--outcome
pairs jointly, capturing richer pairwise interactions and producing a more
precise final ranking. Table~\ref{tab:semantic-retrieval-comparison} summarizes the respective roles, strengths, and limitations of the bi-encoder,
cross-encoder, and resulting hybrid architecture.

\begin{table}[htb]
\centering
\caption{Comparison of bi-encoder, cross-encoder, and hybrid approaches for semantic intent-to-outcome retrieval.}
\label{tab:semantic-retrieval-comparison}
\small
\begin{tabularx}{\textwidth}{
    >{\raggedright\arraybackslash}p{2.3cm}
    >{\raggedright\arraybackslash}X
    >{\raggedright\arraybackslash}p{4.0cm}
    >{\raggedright\arraybackslash}p{3.8cm}
}
\toprule
\textbf{Approach} &
\textbf{Role and Characteristics} &
\textbf{Strengths} &
\textbf{Weaknesses} \\
\midrule

Bi-Encoder &
Used for initial candidate retrieval. Intent and outcome texts are encoded independently, enabling efficient vector-similarity search over large candidate spaces. Model selection was informed by Massive Text Embedding Benchmark (MTEB) sentence-similarity benchmarks~\cite{muennighoff2022mteb}. &
Fast, scalable, and allows outcome embeddings to be precomputed and reused. &
Lower ranking precision because intent--outcome interactions are not evaluated jointly. \\

\addlinespace

Cross-Encoder &
Applied only to shortlisted candidates. Each intent--outcome pair is evaluated jointly, allowing richer semantic interaction and more accurate ranking. &
High semantic accuracy and stronger pairwise ranking. &
More computationally expensive because each candidate pair must be evaluated separately. \\

\addlinespace

Hybrid (chosen) &
Uses the bi-encoder for high-recall candidate retrieval and the cross-encoder for final re-ranking of the shortlisted candidates. &
Combines efficient retrieval with high-precision semantic ranking. &
Introduces slightly greater pipeline complexity than either model used independently. \\

\bottomrule
\end{tabularx}
\end{table}

The hybrid design therefore concentrates the more computationally expensive
pairwise reasoning where it provides the most value, following the same
efficient-retrieval-then-precise-rerank pattern studied by Geigle
et al.~\cite{geigle2022retrievererank} and building on the sentence-encoder
paradigm of Reimers and Gurevych~\cite{reimers2019sentencebert}.

Candidate selection is treated as a ranking problem rather than as probability estimation. Neither cosine similarity nor the cross-encoder score is assumed to be a calibrated probability, making a single fixed cutoff brittle across products and candidate distributions. Instead, the ranked bi-encoder similarities are first examined for changes between consecutive candidates. A sufficiently large score drop provides evidence that semantic relevance is beginning to degrade and can therefore be used to restrict the candidate set before final ranking.

\begin{figure}[htb]
\centering
\includegraphics[width=0.92\textwidth]{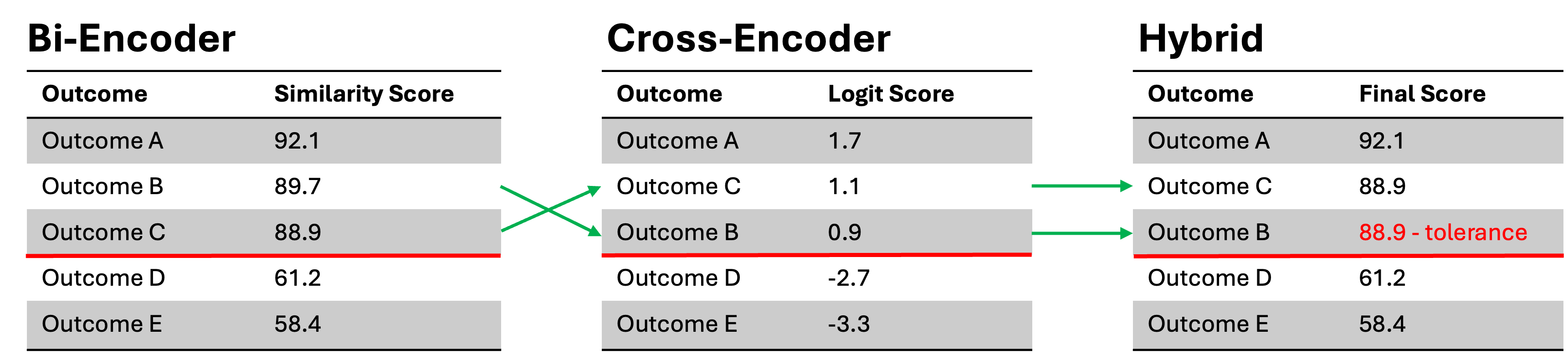}
\caption{Illustration of hybrid ranking and candidate-selection logic. Bi-encoder similarity provides efficient candidate retrieval and score-gap information used to trim the candidate set before reranking, while the cross-encoder determines the final semantic ranking of the shortlisted outcomes. Table~\ref{tab:scoring-thresholding} details the governing parameters.}
\label{fig:hybrid-scoring}
\end{figure}

As shown in Figure~\ref{fig:hybrid-scoring}, the cross-encoder can reorder the shortlisted outcomes when pairwise semantic evaluation identifies a more plausible ranking. The final output ordering follows this cross-encoder ranking, while the displayed relative similarity values originate from the bi-encoder and are adjusted, where necessary, to remain monotonically decreasing with the final rank. The number of returned outcomes is also bounded to prevent noisy long tails; in the current implementation, no more than five business outcomes are returned.

\begin{table}[htb]
\centering
\caption{Scoring and thresholding rules applied during intent-to-outcome candidate selection and ranking.}
\label{tab:scoring-thresholding}
\begin{tabular}{L{3.8cm}L{12.5cm}}
\toprule
\textbf{Mechanism} & \textbf{Rule} \\
\midrule
Score-gap detection & Ranked bi-encoder similarities are scanned for a sufficiently large drop between consecutive candidates; a detected drop is treated as evidence that semantic relevance is beginning to degrade and is used to restrict the candidate set before reranking, rather than relying on a single fixed similarity cutoff. \\
Maximum outcomes returned & No more than five business outcomes are returned per customer intent, bounding the output to prevent noisy long tails. \\
Final ordering & Determined by the cross-encoder ranking over the shortlisted candidates, not by the bi-encoder similarity used for initial retrieval. \\
Displayed similarity scores & Derived from the bi-encoder scores and adjusted, where necessary, to decrease monotonically with the final cross-encoder rank, so that displayed values remain consistent with the output ordering. \\
\bottomrule
\end{tabular}
\end{table}

The hybrid design was selected empirically against bi-encoder-only, cross-encoder-only, and zero-shot classification alternatives. Table~\ref{tab:intent-outcome-evaluation} summarizes the overall performance on the SME-annotated evaluation set using F1 at rank 5 (F1@5) and Mean Reciprocal Rank (MRR) as the primary ranking-quality metrics, alongside latency.

\begin{table}[htb]
\centering
\caption{Overall comparison of intent-to-outcome mapping approaches.
Higher is better for F1 and MRR; lower is better for latency.}
\label{tab:intent-outcome-evaluation}
\begin{tabular}{lccc}
\hline
\textbf{Approach} & \textbf{F1@5} & \textbf{MRR} & \textbf{Latency (s)} \\
\hline
Bi-encoder            & 51.09 & 0.76 & 0.056 \\
Cross-encoder         & \textbf{54.82} & \textbf{0.87} & 0.098 \\
Hybrid                & \textbf{54.82} & \textbf{0.87} & 0.060 \\
Zero-shot classifier  & 47.00 & 0.85 & 1.213 \\
\hline
\end{tabular}
\end{table}

The hybrid pipeline matched the strongest F1@5 and MRR performance, with inference latency close to the bi-encoder-only approach. The cross-encoder-only configuration produced similar ranking quality at higher latency, while the zero-shot classifier was substantially slower. Because the SME annotations were not exhaustive, these measurements are used comparatively to guide architecture selection rather than as an absolute estimate of production accuracy.

The resulting application illustrates the broader value of the intent intelligence layer: the same normalized objective that supports consolidation and analysis can also function directly as a query for downstream retrieval, recommendation, and decision-support systems.

\section{Evaluation and Feedback Loops}

CIE treats evaluation as part of the architecture rather than as a final acceptance test. Because the pipeline contains deterministic transformations, semantic models, generative components, and human-facing outputs, no single metric is sufficient. Evaluation therefore combines repeatability analysis, automated model-based evaluation, and SME review, with different forms of evidence applied to different stages of the system.

For multimodal E2E extraction, one concern is whether repeated executions produce a stable representation of the same source material. A repeatability experiment was conducted by generating the predicted structured output for the same sampled E2E plan across five runs and comparing the extracted table columns and cells using embedding similarity. The experiment produced a mean similarity of $95.19\%$ with a standard deviation of $1.12\%$, with individual runs ranging from $94.07\%$ to $96.31\%$. These results indicate relatively low run-to-run variation for the sampled document. Importantly, repeatability measures stability rather than correctness: a consistently reproduced output may still omit or misinterpret source information, so this measure is considered alongside ground-truth and SME-based evaluation.

Automatic evaluation was also applied to intent extraction from E2E-sourced evidence, using an independent LLM as a judge, following the LLM-as-a-judge methodology of Zheng et al.~\cite{zheng2023judging}. This targets a different stage of the pipeline than the repeatability experiment above: rather than the fidelity of structured context recovery, it assesses the candidate intents generated from that context. Gemini 2.5 Flash~\cite{geminiteam2025gemini25} was used as the evaluator at temperature zero, with each metric averaged across ten runs. Table~\ref{tab:e2e-llm-evaluation} summarizes the reported results.

\begin{table}[htb]
\centering
\caption{Automatic evaluation of intent extraction from E2E-sourced evidence, using an independent LLM judge. Values are the mean and standard deviation reported across ten evaluation runs.}
\label{tab:e2e-llm-evaluation}
\begin{tabular}{L{4.0cm}L{3.0cm}L{3.0cm}L{6.0cm}}
\toprule
\textbf{Metric} & \textbf{Mean} & \textbf{Standard deviation} & \textbf{Interpretation} \\
\midrule
Faithfulness & 93.5\% & 7.9\% &
Consistency of the generated response with the retrieved source context. \\

Faithfulness vs.\ ground truth & 66.1\% & 31.9\% &
Faithfulness when evaluated relative to the manually established reference output. \\

Semantic similarity & 48.5 & 10.7 &
Semantic resemblance between the generated output and the reference answer. \\

Answer correctness & 37.4\% & 14.4\% &
Combined assessment of semantic and factual agreement with the reference answer. \\
\bottomrule
\end{tabular}
\end{table}

The difference between high context faithfulness and the lower ground-truth-based metrics is informative. It suggests that an output can remain well grounded in the context supplied to the model while still differing materially from a manually defined reference representation. In document-processing pipelines, this distinction is useful because errors can originate either in the generation stage itself or earlier, through incomplete or imperfect extraction of the source context. The relatively high variance of the ground-truth faithfulness measure also reinforces the need to consider aggregate automatic scores together with example-level inspection rather than as a single acceptance threshold.

\enlargethispage{\baselineskip}
Human evaluation provides an additional layer of evidence. During a pilot evaluation of intent extraction, SMEs reviewed inferred intents for five customers and supplied positive or negative feedback at the intent level. Across these customers, average intent-extraction accuracy was $78\%$. A separate outcome-benchmarking evaluation collected feedback for 20 customers spanning four products. SMEs judged all recommended business outcomes to be relevant to their respective customers, while $81.6\%$ of the selected peers were judged relevant. The lower peer-relevancy result was attributed primarily to occasional inaccuracies in the customer industry information used during peer grouping. Table~\ref{tab:sme-evaluation} summarizes these results.

\begin{table}[htb]
\centering
\caption{Summary of SME-based evaluation results.}
\label{tab:sme-evaluation}
\begin{tabular}{L{3.0cm}L{3.0cm}L{3.5cm}L{6.5cm}}
\toprule
\textbf{Evaluation target} & \textbf{Evaluation scope} & \textbf{Result} & \textbf{Interpretation} \\
\midrule
Intent extraction &
Five customers &
78\% average accuracy &
SMEs assessed whether individual inferred intents were supported and relevant for each customer. \\

Outcome relevancy &
20 customers across four products &
100\% &
SMEs confirmed that all recommended business outcomes were relevant to the respective customers. \\

Peer relevancy &
20 customers across four products &
81.6\% &
Most selected peers were judged relevant; remaining errors were associated primarily with inaccuracies in customer industry metadata used for peer grouping. \\
\bottomrule
\end{tabular}
\end{table}

These evaluations reflect the broader principle that different stages require different forms of evidence. Multimodal context extraction can be evaluated against manually validated structured ground truth using cell/key similarity, edit distance, structural correctness, and repeatability. Intent extraction can be assessed for faithfulness to retrieved context, semantic and factual correctness against SME-validated references, and stability across repeated runs. Post-processing requires additional cluster-focused measures such as purity, singleton rate, cluster count, outlier reassignment acceptance, keyword coverage and distinctiveness, aggregation faithfulness, and information preservation. Downstream applications introduce further task-specific measures, such as peer relevance for MGI and ranking quality for intent-to-outcome retrieval.

The development process also showed that SME feedback itself requires controls. Directly written SME intents did not always align with the source artifact, product scope, or the operational intent guidelines. A later validation round, in which SME-provided statements were rewritten according to the intent guidelines and then reviewed, produced substantially better qualitative agreement. This motivates a feedback design in which SMEs validate evidence-grounded candidate representations rather than acting as an assumed infallible source of truth.

Taken together, these evaluation layers make feedback part of the system's iterative development loop. Automated metrics provide repeatable signals at scale, repeatability testing exposes instability in stochastic components, and SME review tests whether technically plausible outputs remain meaningful in the business context. Evaluation is therefore used not only to quantify final performance, but also to identify which component of the pipeline should be refined when failures occur.

\section{Discussion}

Three engineering lessons emerge from the work. First, customer intent extraction is primarily a data and representation problem rather than only a model-selection problem. Large models can produce fluent candidates, but much of the practical quality improvement comes from source-specific filtering, structured context recovery, canonical metadata, semantic grouping, and constrained aggregation. The post-processing representation determines whether model outputs are reusable.

Second, deterministic and probabilistic techniques should be assigned to the problems they solve best. Regex filtering, exact deduplication, and canonical mappings handle explicit data-quality rules. Embeddings and clustering handle semantic relatedness. LLMs are most useful where language understanding, semantic descriptor generation, or constrained synthesis is required. Treating every transformation as an LLM call would be harder to debug, more expensive, and less reproducible.

Third, ambiguity should remain visible. A singleton is not automatically an error; it may encode a rare but legitimate customer objective. Similarly, a short candidate may be ambiguous in isolation but meaningful when paired with known product or domain metadata. The architecture therefore adds context for semantic computation without overwriting the original text, and it allows uncertain items to remain independent.

More broadly, the system demonstrates a reusable pattern for LLM-based information extraction beyond customer intent. Any workflow that deliberately extracts a high-recall set of candidate insights from large heterogeneous corpora can face the same downstream problem: semantic duplication, inconsistent granularity, noise, and an output volume that is difficult to consume. A reduction layer that preserves provenance and uncertainty can therefore be treated as a first-class architectural component rather than a final formatting step~\cite{aminiranjbar2025dawn}.

\section{Limitations}

The current article describes a system architecture and development evidence rather than a completed controlled benchmark.

The adaptive clustering thresholds were selected as conservative operational starting points and are not universal boundaries. They should be recalibrated for new embedding models, candidate-set distributions, and business domains. The large-set path also inherits sensitivity from dimensionality reduction and density-based clustering, while the small-set graph path can exhibit chaining when similarity thresholds are too permissive.

Keyword generation and constrained aggregation still depend on an LLM. Although the prompt explicitly restricts generation to the information present in cluster members, factuality must be guardrailed and evaluated rather than assumed. Similarly, metadata padding can improve representation of underspecified text but may also overemphasize a product or domain label if the metadata itself is incorrect.

Finally, the article currently focuses on English-language enterprise data and on customer-intent use cases within the described environment. Broader validation is required before claiming that the same thresholds, prompts, or clustering choices transfer unchanged to other domains.

\section{Conclusion}

Customer intent extraction is not a single inference call. It is a controlled transformation from heterogeneous customer evidence to reusable business intelligence. The architecture described in this paper deliberately separates high-recall extraction from high-utility post-processing: it first preserves potentially relevant objectives, then applies deterministic normalization, metadata-aware semantic representation, adaptive clustering, explainable keyword generation, conservative outlier handling, and constrained aggregation.

The resulting stable intent unit is both concise and auditable. Downstream systems receive a normalized statement, while the underlying member intents, source references, clustering decisions, keywords, and model versions remain available for explanation and reprocessing. This design makes semantic reduction a first-class component of enterprise LLM pipelines and provides a foundation for predictive intent generation, intent-guided semantic retrieval and mapping, and quality evaluation.

\bibliographystyle{ieeetr}
\bibliography{references_v3}

@inproceedings{reimers2019sentencebert,
  author    = {Reimers, Nils and Gurevych, Iryna},
  title     = {Sentence-{BERT}: Sentence Embeddings using {S}iamese {BERT}-Networks},
  booktitle = {Proceedings of the 2019 Conference on Empirical Methods in Natural Language Processing (EMNLP)},
  year      = {2019},
  eprint    = {1908.10084},
  archivePrefix = {arXiv},
  primaryClass  = {cs.CL},
  url       = {https://arxiv.org/abs/1908.10084}
}

@article{mcinnes2018umap,
  author  = {McInnes, Leland and Healy, John and Melville, James},
  title   = {{UMAP}: Uniform Manifold Approximation and Projection for Dimension Reduction},
  year    = {2018},
  eprint  = {1802.03426},
  archivePrefix = {arXiv},
  primaryClass  = {stat.ML},
  url     = {https://arxiv.org/abs/1802.03426}
}

@article{mcinnes2017hdbscan,
  author  = {McInnes, Leland and Healy, John},
  title   = {Accelerated Hierarchical Density Clustering},
  year    = {2017},
  eprint  = {1705.07321},
  archivePrefix = {arXiv},
  primaryClass  = {stat.ML},
  url     = {https://arxiv.org/abs/1705.07321}
}

@article{huang2023hallucination,
  author  = {Huang, Lei and Yu, Weijiang and Ma, Weitao and Zhong, Weihong and Feng, Zhangyin and Wang, Haotian and Chen, Qianglong and Peng, Weihua and Feng, Xiaocheng and Qin, Bing and Liu, Ting},
  title   = {A Survey on Hallucination in Large Language Models: Principles, Taxonomy, Challenges, and Open Questions},
  year    = {2023},
  eprint  = {2311.05232},
  archivePrefix = {arXiv},
  primaryClass  = {cs.CL},
  url     = {https://arxiv.org/abs/2311.05232}
}

@inproceedings{zheng2023judging,
  author    = {Zheng, Lianmin and Chiang, Wei-Lin and Sheng, Ying and Zhuang, Siyuan and Wu, Zhanghao and Zhuang, Yonghao and Lin, Zi and Li, Zhuohan and Li, Dacheng and Xing, Eric P. and Zhang, Hao and Gonzalez, Joseph E. and Stoica, Ion},
  title     = {Judging {LLM}-as-a-Judge with {MT}-Bench and Chatbot Arena},
  booktitle = {Advances in Neural Information Processing Systems (NeurIPS)},
  year      = {2023},
  eprint    = {2306.05685},
  archivePrefix = {arXiv},
  primaryClass  = {cs.CL},
  url       = {https://arxiv.org/abs/2306.05685}
}

@article{abbas2023semdedup,
  author  = {Abbas, Amro and Tirumala, Kushal and Simig, D{\'a}niel and Ganguli, Surya and Morcos, Ari S.},
  title   = {{SemDeDup}: Data-efficient learning at web-scale through semantic deduplication},
  year    = {2023},
  eprint  = {2303.09540},
  archivePrefix = {arXiv},
  primaryClass  = {cs.LG},
  url     = {https://arxiv.org/abs/2303.09540}
}

@article{geigle2022retrievererank,
  author  = {Geigle, Gregor and Pfeiffer, Jonas and Reimers, Nils and Vuli{\'c}, Ivan and Gurevych, Iryna},
  title   = {Retrieve Fast, Rerank Smart: Cooperative and Joint Approaches for Improved Cross-Modal Retrieval},
  journal = {Transactions of the Association for Computational Linguistics (TACL)},
  year    = {2022},
  eprint  = {2103.11920},
  archivePrefix = {arXiv},
  primaryClass  = {cs.CV},
  url     = {https://arxiv.org/abs/2103.11920}
}

@article{corbeil2022intentlandscape,
  author  = {Corbeil, Jean-Philippe and Li, Mia Taige and Ghavidel, Hadi Abdi},
  title   = {Building the Intent Landscape of Real-World Conversational Corpora with Extractive Question-Answering Transformers},
  year    = {2022},
  eprint  = {2208.12886},
  archivePrefix = {arXiv},
  primaryClass  = {cs.CL},
  url     = {https://arxiv.org/abs/2208.12886}
}

@inproceedings{manias2024intentbased,
  author    = {Manias, Dimitrios Michael and Chouman, Ali and Shami, Abdallah},
  title     = {Towards Intent-Based Network Management: Large Language Models for Intent Extraction in {5G} Core Networks},
  booktitle = {Proceedings of the International Conference on the Design of Reliable Communication Networks (DRCN)},
  year      = {2024},
  eprint    = {2403.02238},
  archivePrefix = {arXiv},
  primaryClass  = {cs.NI},
  url       = {https://arxiv.org/abs/2403.02238}
}

@article{liu2025adretrieval,
  author  = {Liu, Tongtong and Wang, Zhaohui and Qin, Meiyue and Lu, Zenghui and Chen, Xudong and Yang, Yuekui and Shu, Peng},
  title   = {Real-time Ad Retrieval via {LLM}-Generative Commercial Intention for Sponsored Search Advertising},
  year    = {2025},
  eprint  = {2504.01304},
  archivePrefix = {arXiv},
  primaryClass  = {cs.IR},
  url     = {https://arxiv.org/abs/2504.01304}
}

@article{cheng2025meituan,
  author  = {Cheng, Xuxin and Zeng, Ke and Cao, Zhiquan and Dai, Linyi and Gao, Wenxuan and Han, Fei and Jian, Ai and Hong, Feng and Hu, Wenxing and Huang, Zihe and Kong, Dejian and Leng, Jia and Liao, Zhuoyuan and Liu, Pei and Lin, Jiaye and Ma, Xing and Ruan, Jingqing and Song, Jiaxing and Tan, Xiaoyu and Xiao, Ruixuan and Yu, Wenhui and Zhan, Wenyu and Zhang, Haoxing and Zhou, Chao and Zhou, Hao and others},
  title   = {Higher Satisfaction, Lower Cost: A Technical Report on How {LLM}s Revolutionize {M}eituan's Intelligent Interaction Systems},
  year    = {2025},
  eprint  = {2510.13291},
  archivePrefix = {arXiv},
  primaryClass  = {cs.CL},
  url     = {https://arxiv.org/abs/2510.13291}
}

@article{viljanen2020coldstart,
  author  = {Viljanen, Markus and Vahlo, Jukka and Koponen, Aki and Pahikkala, Tapio},
  title   = {Content Based Player and Game Interaction Model for Game Recommendation in the Cold Start Setting},
  year    = {2020},
  eprint  = {2009.08947},
  archivePrefix = {arXiv},
  primaryClass  = {cs.IR},
  url     = {https://arxiv.org/abs/2009.08947}
}

@article{aminiranjbar2025dawn,
  author  = {Aminiranjbar, Zahra and Tang, Jianan and Wang, Qiudan and Pant, Shubha and Viswanathan, Mahesh},
  title   = {{DAWN}: Designing Distributed Agents in a Worldwide Network},
  journal = {IEEE Access},
  volume  = {13},
  year    = {2025},
  doi     = {10.1109/ACCESS.2025.3588425}
}

@inproceedings{muennighoff2022mteb,
  author    = {Muennighoff, Niklas and Tazi, Nouamane and Magne, Lo{\"i}c and Reimers, Nils},
  title     = {{MTEB}: Massive Text Embedding Benchmark},
  booktitle = {Proceedings of the 17th Conference of the European Chapter of the Association for Computational Linguistics (EACL)},
  year      = {2023},
  eprint    = {2210.07316},
  archivePrefix = {arXiv},
  primaryClass  = {cs.CL},
  url       = {https://arxiv.org/abs/2210.07316}
}

@article{geminiteam2025gemini25,
  author  = {{Gemini Team}},
  title   = {Gemini 2.5: Pushing the Frontier with Advanced Reasoning, Multimodality, Long Context, and Next Generation Agentic Capabilities},
  year    = {2025},
  eprint  = {2507.06261},
  archivePrefix = {arXiv},
  primaryClass  = {cs.CL},
  url     = {https://arxiv.org/abs/2507.06261}
}

\end{document}